\documentclass[conference]{IEEEtran}
\IEEEoverridecommandlockouts

\usepackage{cite}
\usepackage{amsmath,amssymb,amsfonts}
\usepackage{algorithmic}
\usepackage{graphicx}
\usepackage{textcomp}
\usepackage{xcolor}
\def\BibTeX{{\rm B\kern-.05em{\sc i\kern-.025em b}\kern-.08em
    T\kern-.1667em\lower.7ex\hbox{E}\kern-.125emX}}
\begin{document}

\title{Toward Personalized Digital Twins for Cognitive Decline Assessment: A Multimodal, Uncertainty-Aware Framework}

\author{\IEEEauthorblockN{Bulent Soykan}
\IEEEauthorblockA{\textit{Department of Mechanical, Industrial \& Manufac. Eng.} \\
\textit{The University of Toledo}\\
Toledo, OH, USA \\
Bulent.Soykan@UToledo.Edu}
\and
\IEEEauthorblockN{Gulsah Hancerliogullari Koksalmis}
\IEEEauthorblockA{\textit{Department of Industrial Eng. and Mngt. Systems} \\
\textit{University of Central Florida}\\
Orlando, FL, USA \\
gulsah.hancerliogullarikoksalmis@ucf.edu}
\and
\IEEEauthorblockN{Hsin-Hsiung Huang}
\IEEEauthorblockA{\textit{Department of Statistics and Data Science} \\
\textit{University of Central Florida}\\
Orlando, FL, USA \\
hsin-hsiung.huang@ucf.edu}
\and
\IEEEauthorblockN{Laura J. Brattain}
\IEEEauthorblockA{\textit{Department of Internal Medicine} \\
\textit{University of Central Florida}\\
Orlando, FL, USA \\
laura.brattain@ucf.edu}
}

\maketitle

\begin{abstract}
Cognitive decline is highly heterogeneous across individuals, which complicates prognosis, trial design, and treatment planning. We present the Personalized Cognitive Decline Assessment Digital Twin (PCD-DT), a multimodal and uncertainty-aware framework for modeling patient-specific disease trajectories from sparse, noisy, and irregular longitudinal data. The framework combines three methodological components: (1) latent state-space models for individualized temporal dynamics, (2) multimodal fusion for clinical, biomarker, and imaging features, and (3) uncertainty-aware validation and adaptive updating for robust digital twin operation. We also outline how conditional generative models can support data augmentation and stress testing for underrepresented progression patterns. As a preliminary feasibility study, we analyze longitudinal TADPOLE trajectories and show clear separation between cognitively normal and Alzheimer's disease cohorts in ADAS13, ventricle volume, and hippocampal volume over five years. We further conduct a multimodal next-visit prediction ablation using an LSTM sequence model on 3{,}003 visit-pair sequences derived from TADPOLE, where the combined cognitive plus MRI configuration achieves the lowest standardized RMSE for both ADAS13 (0.4419) and ventricle volume (0.5842), outperforming a Last Observation Carried Forward baseline. A Bayesian tensor modeling component for high-dimensional imaging fusion is also discussed. These results support the feasibility of the proposed architecture while also highlighting the need for stronger uncertainty calibration and longer-horizon predictive evaluation. The PCD-DT framework provides a principled starting point for personalized in silico modeling in neurodegenerative disease. This work positions PCD-DT as a foundational step toward clinically deployable, uncertainty-aware digital twin systems.
\end{abstract}


\section{Introduction}
Cognitive decline (CD), including Alzheimer's disease (AD), remains a major clinical and public-health challenge because patients differ substantially in age of onset, rate of progression, and symptom profile \cite{javaid2021epidemiology, nandi2022global}. This heterogeneity limits the usefulness of population-level averages for prognosis and complicates the design of clinical trials and personalized interventions.

Digital twins (DTs) offer a compelling direction for this problem by maintaining a virtual, data-driven representation of an individual that can be updated over time \cite{NAP26894}. For cognitive decline, such a twin can combine multimodal observations across visits, estimate latent disease state, forecast future outcomes, and quantify uncertainty around those forecasts. However, realizing this vision remains difficult because available data are sparse, noisy, irregularly sampled, and incomplete across modalities. Current methods also struggle to connect prediction with clinically interpretable uncertainty, which is essential for trustworthy deployment \cite{2025StatisticalLearning}.

This paper introduces the Personalized Cognitive Decline Assessment Digital Twin (PCD-DT), a framework that integrates latent state-space modeling, multimodal fusion, synthetic data augmentation, and uncertainty-aware validation. The main contributions are threefold. First, it defines an architecture for patient-specific DT updating from longitudinal multimodal data. Second, it outlines a modeling stack that combines deep state-space methods with Bayesian dimensionality reduction for parsimonious fusion. Third, it reports preliminary feasibility analyses showing that longitudinal biomarkers in TADPOLE contain strong disease-progression signal. We intentionally frame the current work as a foundation for an end-to-end DT rather than a completed clinical decision tool. Unlike prior approaches, the proposed framework explicitly integrates temporal dynamics, multimodal fusion, and uncertainty-aware updating within a single unified architecture.

\section{Related Work}
Deep learning has advanced the early detection and classification of AD from MRI, PET, cognitive assessments, and other heterogeneous data sources \cite{nagarajan2025comprehensive, kale2024ai, lee2019predicting, garcia2024predicting}. Recent multimodal architectures combining convolutional neural networks and transformer-based models have shown strong performance in Alzheimer’s disease prediction by capturing complex cross-modal relationships \cite{chen2025multimodal, yao2025interpretable, guo2025meta}. Despite these advances, many approaches remain focused on static or short-horizon prediction rather than continuously updated patient representations. 

Longitudinal modeling has been approached with recurrent neural networks and state-space models. RNN-based approaches have shown strong performance for forecasting future cognitive status from visit histories \cite{nguyen2020predicting}. State-space models are well suited for clinical trajectories because they explicitly separate latent disease state from noisy observations and naturally support filtering and forecasting \cite{chua2022state, burkhart2024unsupervised}. However, integrating these temporal models with multimodal data in a unified and uncertainty-aware manner remains an open challenge.

Digital twins in medicine aim to unify these advances into patient-specific computational models that evolve over time with data \cite{sprint2024building, ashraf2024digital}. Recent studies have explored digital twin frameworks for cross-modal patient modeling and clinical trial optimization using synthetic cohorts \cite{amato2026digital, wang2025using}. Generative approaches have also been explored for simulating AD progression and for creating virtual control populations \cite{fisher2018deep, bertolini2021forecasting, bertolini2020modeling}. Our work builds on this literature by explicitly integrating multimodal fusion, temporal modeling, and uncertainty-aware updating within a single unified architecture. Unlike prior approaches, which typically address these components in isolation, the proposed framework provides a cohesive and extensible foundation for personalized digital twin systems.

\section{PCD-DT Framework}
\subsection{System Overview}
The PCD-DT architecture links four components in a closed loop: (1) data ingestion and temporal state estimation, (2) multimodal representation learning, (3) synthetic-data generation for augmentation and stress testing, and (4) verification, validation, and uncertainty quantification (VVUQ). Figure~\ref{fig:architecture} summarizes this workflow. This design enables continuous, patient-specific updating and distinguishes the framework from static or batch prediction approaches.

\begin{figure*}[htbp]
    \centering
    \includegraphics[width=0.9\textwidth]{figures/Figure1.png}
    \caption{Overview of the proposed PCD-DT framework. The architecture links multimodal inputs, temporal state estimation, multimodal fusion, synthetic-data generation, and VVUQ within a continuously updated patient-specific digital twin.}
    \label{fig:architecture}
\end{figure*}

\subsection{Data and Update Loop}
We use the TADPOLE longitudinal dataset as the primary framework test bed. The data ingestion layer standardizes multimodal inputs including MRI and PET summary measures, cerebrospinal fluid biomarkers, cognitive scores such as ADAS13 and MMSE, demographic variables, and diagnosis labels. These measurements are irregular in time and contain missing values, so the twin represents patient condition through a latent state $z_t$ and treats the observed visit data $x_t$ as noisy partial evidence.

As new observations arrive, the twin updates the posterior over latent state and produces revised forecasts. Anomaly-detection logic is attached to this update loop so that large deviations between predicted and observed trajectories can trigger focused posterior updating, re-evaluation, or clinician review.

\section{Methodological Components}
\subsection{State-Space Modeling for Individualized Dynamics}
We model disease progression with a latent state-space formulation,
\begin{align}
    z_t &\sim p_\theta(z_t\mid z_{t-1}), \\
    x_t &\sim p_\phi(x_t\mid z_t),
\end{align}
where $z_t \in \mathbb{R}^d$ denotes latent disease state and $x_t \in \mathbb{R}^m$ denotes multimodal clinical observations. This separation is particularly suited to cognitive-decline data because it filters measurement noise, accommodates irregular visit timing, and supports forecasting.

To capture nonlinear dynamics, the transition and emission models are parameterized with recurrent or deep Markov architectures. Because exact posterior inference is generally intractable, we use variational inference with an approximate posterior $q(z_t\mid x_{\le t}, z_{<t})$. In practice, masking and model-based imputation are used to handle missing entries, and continuous-time variants can be used when visit spacing varies substantially. This formulation provides a natural foundation for individualized forecasting and uncertainty-aware trajectory estimation.

\subsection{Multimodal Fusion and Dimensionality Reduction}
The framework combines deep fusion and Bayesian parsimony. Transformer-style attention can learn context-dependent weights across modalities and time. These components produce a fused representation $R_t$ that is more informative than simple feature concatenation.

For high-dimensional imaging inputs, we use Bayesian dimensionality reduction to preserve interpretability and computational tractability. Sparse priors such as spike-and-slab or horseshoe formulations can identify informative regions of interest. For structured spatiotemporal inputs, a regularized low-rank tensor representation can be written as
\begin{equation}
    \mathbf{C}_k = \sum_{r=1}^{R} \mathbf{A}_{rk} \otimes \mathbf{B}_{rk},
\end{equation}
where shrinkage priors support adaptive rank selection and uncertainty-aware coefficient estimation. This approach supports both predictive performance and interpretability, which are critical for clinical adoption.

\subsection{Synthetic Augmentation and VVUQ}
Class imbalance and limited follow-up remain serious constraints in dementia datasets. To address this, the framework includes conditional VAEs and GANs for generating targeted synthetic trajectories conditioned on baseline phenotype, demographics, or genotype. In the current paper we describe this as a framework component rather than a fully validated module. This enables targeted augmentation of underrepresented disease progression patterns, improving robustness and fairness.

Uncertainty quantification is built into both state estimation and prediction by propagating posterior uncertainty to future observations. The VVUQ layer also includes anomaly detection and fairness checks across demographic subgroups, which are necessary for trustworthy clinical deployment.

\section{Preliminary Results}
\subsection{Longitudinal Trajectory Separation in TADPOLE}
\begin{figure*}[htbp]
    \centering
    \includegraphics[width=\textwidth]{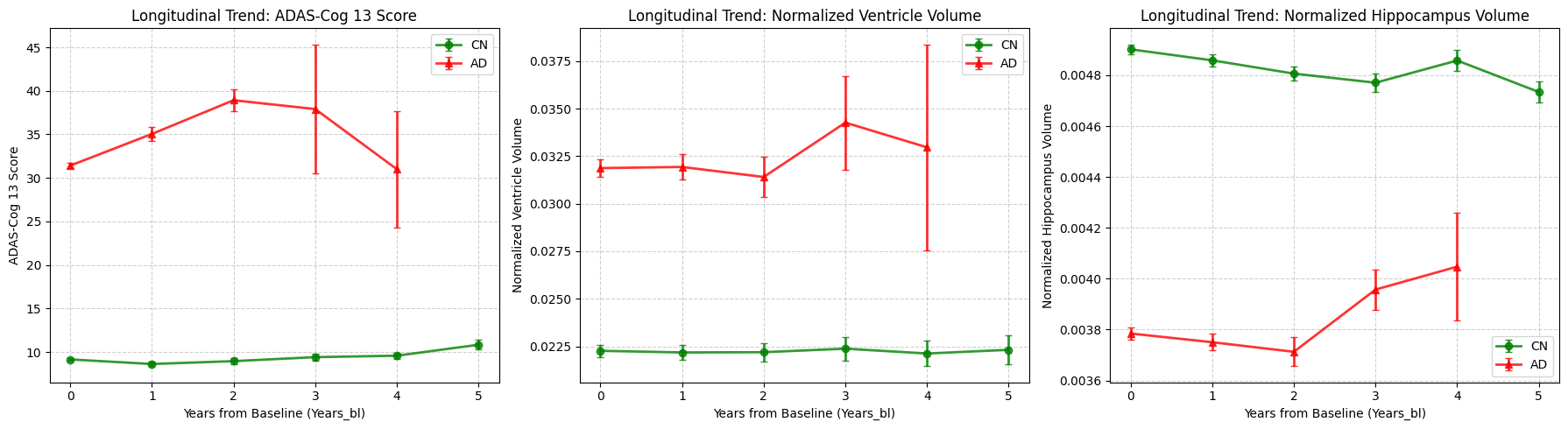}
    \caption{Average longitudinal trends for key biomarkers in TADPOLE. Plots show mean $\pm$ standard error for ADAS13, normalized ventricle volume, and normalized hippocampal volume over five years from baseline for cognitively normal (CN) and Alzheimer's disease (AD) groups.}
    \label{fig:longitudinal_trends}
\end{figure*}

Our preliminary feasibility analysis uses curated TADPOLE longitudinal records to compare baseline cognitively normal (CN) and AD cohorts over a five-year window. Figure~\ref{fig:longitudinal_trends} demonstrates clear separation between cognitively normal and Alzheimer’s disease groups across key cognitive and structural biomarkers. The AD cohort exhibits consistently worse ADAS13 scores, larger normalized ventricle volume, and smaller normalized hippocampal volume than the CN cohort. These expected patterns indicate that the longitudinal data carry strong progression signal that a patient-specific state-space model can exploit. These findings provide initial evidence that patient-specific digital twin modeling is feasible using real-world longitudinal data.

\subsection{Next-Visit Prediction with Multimodal Ablation}
To evaluate the predictive feasibility of the temporal modeling component, we conducted a next-visit prediction study on TADPOLE using an LSTM-based sequence model with a mixed-target objective: standardized regression for ADAS13 and ventricle volume, and three-class classification for diagnosis (CN, MCI, AD). After median imputation, standardization, and removal of records with missing targets, the curated cohort yielded 3{,}932 cleaned visit records and 3{,}003 valid history-target sequence pairs. We compared four modality configurations against a Last Observation Carried Forward (LOCF) baseline: cognitive features only (CDRSB, ADAS11/13, MMSE, RAVLT, FAQ, MOCA); structural MRI features only (ventricles, hippocampus, whole brain, entorhinal, fusiform, mid-temporal, ICV); cognitive plus MRI; and a full model that adds demographics (age, sex, education, APOE4). The LSTM used two stacked layers, hidden size 64, dropout 0.3, and was trained with Adam (learning rate $10^{-3}$, batch size 32) for 20 epochs on an 80/15/5 train/validation/test split with a fixed random seed.

Table~\ref{tab:ablation} reports standardized RMSE on the held-out test set. The combined cognitive plus MRI configuration achieved the lowest ADAS13 RMSE (0.4419) and ventricle RMSE (0.5842), outperforming the LOCF baseline on both targets and indicating that multimodal fusion captures complementary progression signal beyond either modality alone. Cognitive-only features were weaker for ventricle forecasting, while MRI-only features underperformed for cognitive scoring, consistent with each modality's measurement focus. Adding demographic covariates to the full model did not yield further improvement over the cognitive plus MRI configuration in this short-horizon setting.

\begin{table}[htbp]
\caption{Next-visit prediction performance on TADPOLE (standardized RMSE; lower is better). Best per column in bold.}
\label{tab:ablation}
\centering
\begin{tabular}{lcc}
\hline
\textbf{Model} & \textbf{ADAS13 RMSE} & \textbf{Ventricles RMSE} \\
\hline
LOCF baseline & 0.4612 & 0.6569 \\
LSTM (Cognitive) & 0.4807 & 1.0293 \\
LSTM (MRI) & 0.6344 & 0.5873 \\
LSTM (Cognitive + MRI) & \textbf{0.4419} & \textbf{0.5842} \\
LSTM (Full Model) & 0.4494 & 0.6567 \\
\hline
\end{tabular}
\end{table}

\begin{figure}[!htbp]
    \centering
    \includegraphics[width=0.95\columnwidth]{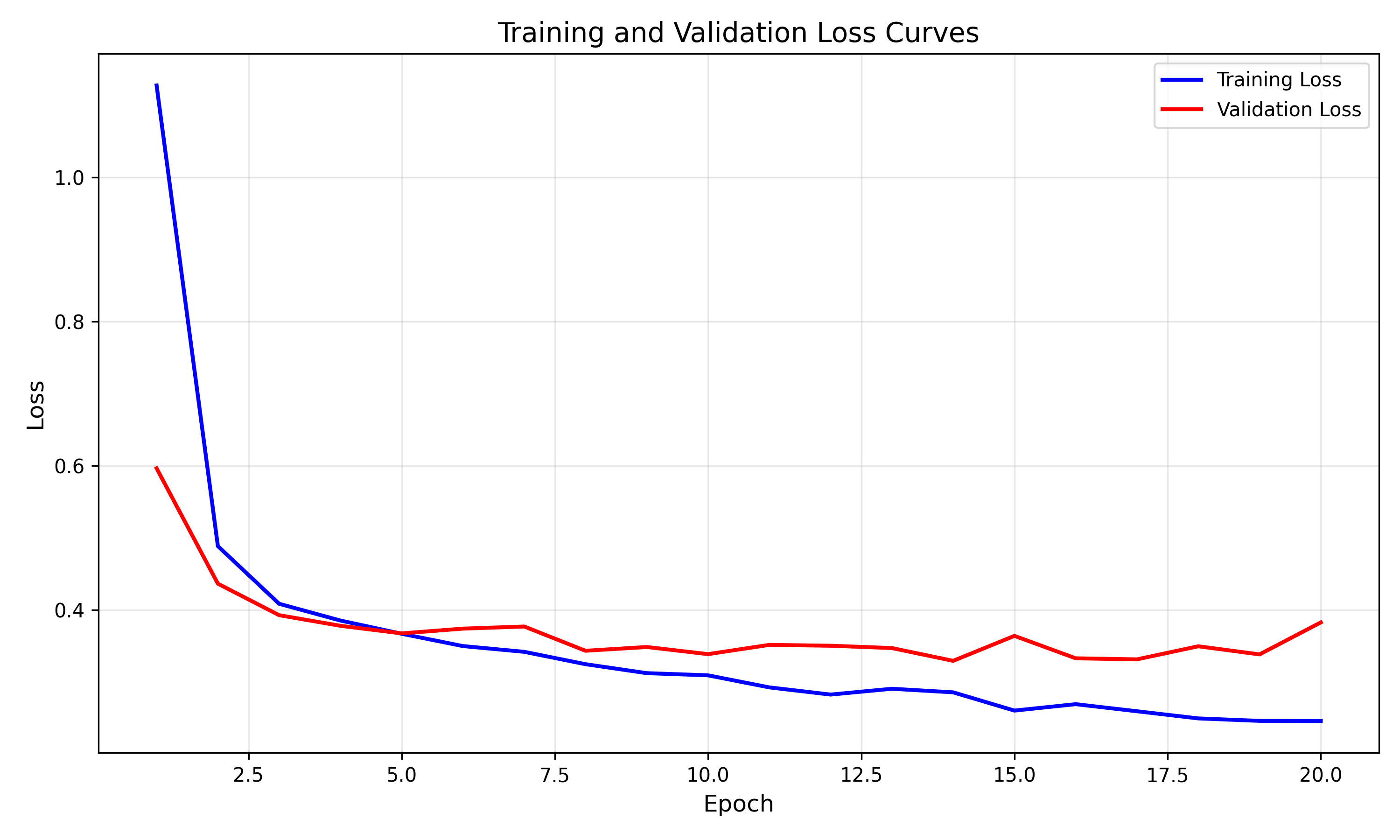}
    \caption{Training and validation loss curves for the LSTM full model on the next-visit prediction task. Smooth convergence and a small generalization gap support the chosen capacity and regularization.}
    \label{fig:training_curves}
\end{figure}

\begin{figure}[!htbp]
    \centering
    \includegraphics[width=0.95\columnwidth]{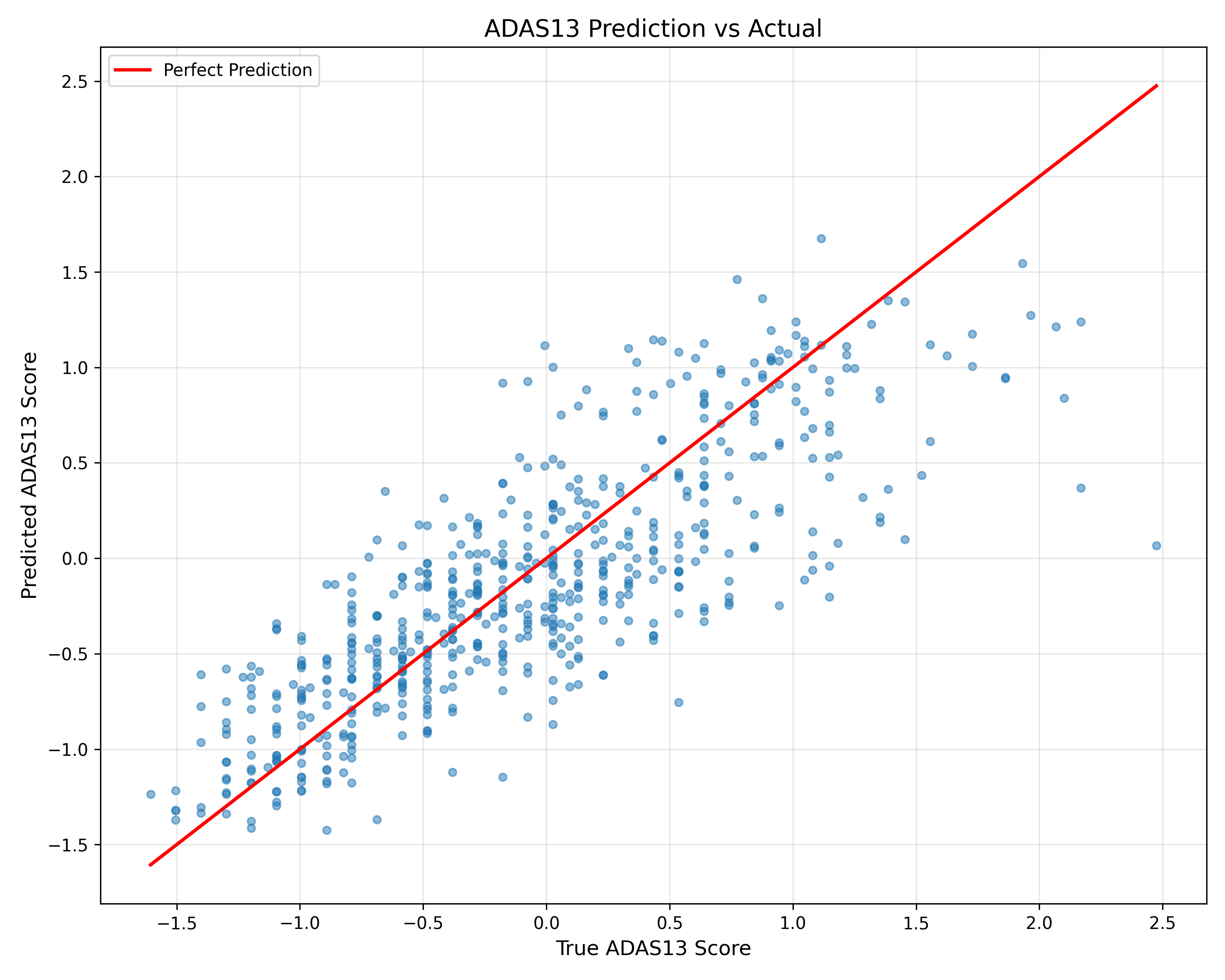}
    \caption{Predicted versus actual standardized ADAS13 scores for the LSTM full model on the held-out test set. The red diagonal denotes perfect prediction; clustering near the identity line across the score range indicates patient-level discrimination beyond mean regression.}
    \label{fig:prediction_scatter}
\end{figure}

Figure~\ref{fig:training_curves} shows training and validation loss curves for the full model, exhibiting smooth convergence and a small generalization gap, consistent with the dropout regularization used. Figure~\ref{fig:prediction_scatter} compares predicted and actual standardized ADAS13 scores on the test set, where points cluster around the identity line across the cognitive range, indicating useful patient-level discrimination rather than mean-regression behavior.

\begin{figure}[!htbp]
    \centering
    \includegraphics[width=0.95\columnwidth]{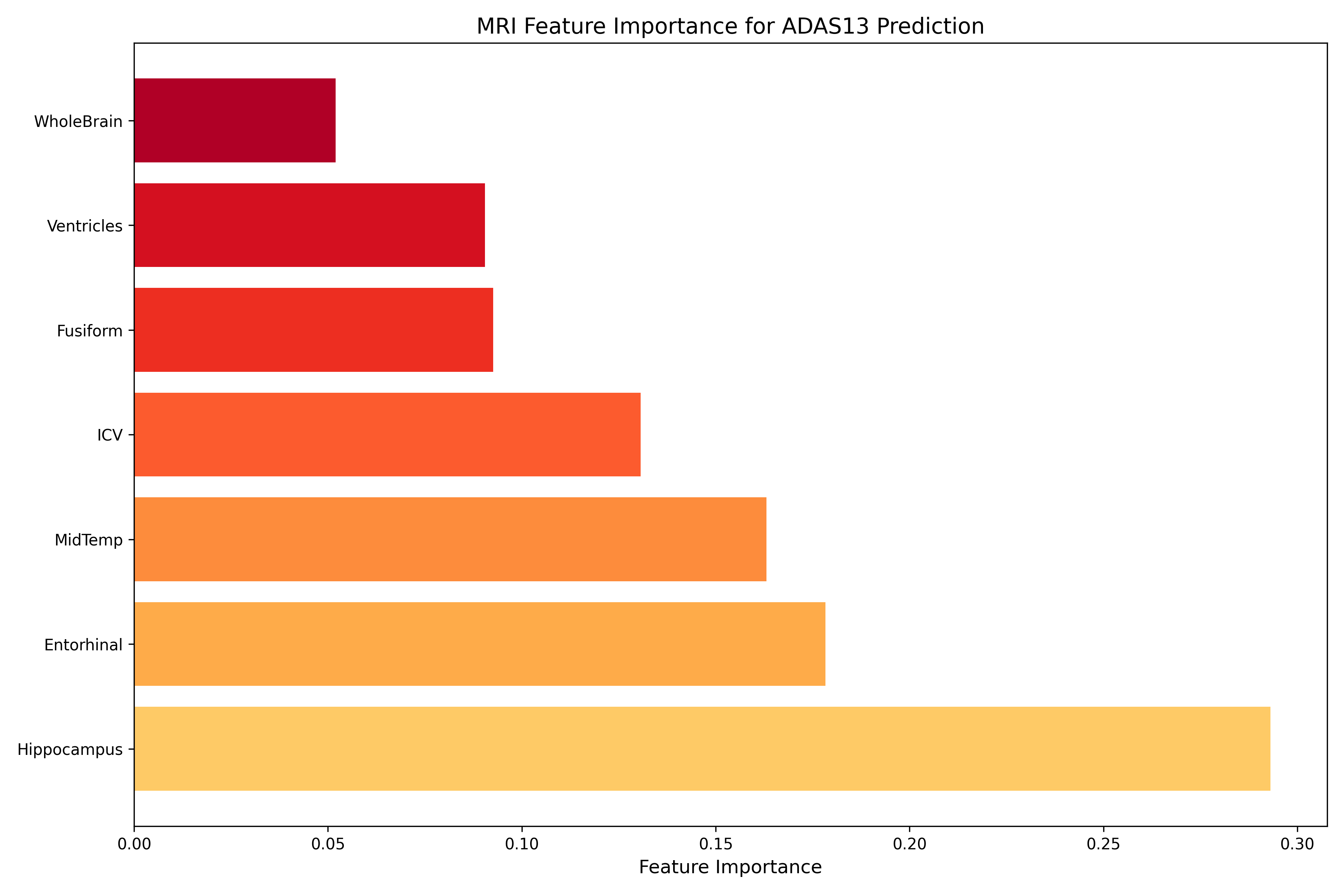}
    \caption{Ranked structural MRI feature importance for ADAS13 next-visit prediction obtained from a gradient-boosted surrogate model. Hippocampal and ventricular measurements dominate, consistent with established AD imaging biomarkers.}
    \label{fig:feature_importance}
\end{figure}

\begin{figure}[!htbp]
    \centering
    \includegraphics[width=0.95\columnwidth]{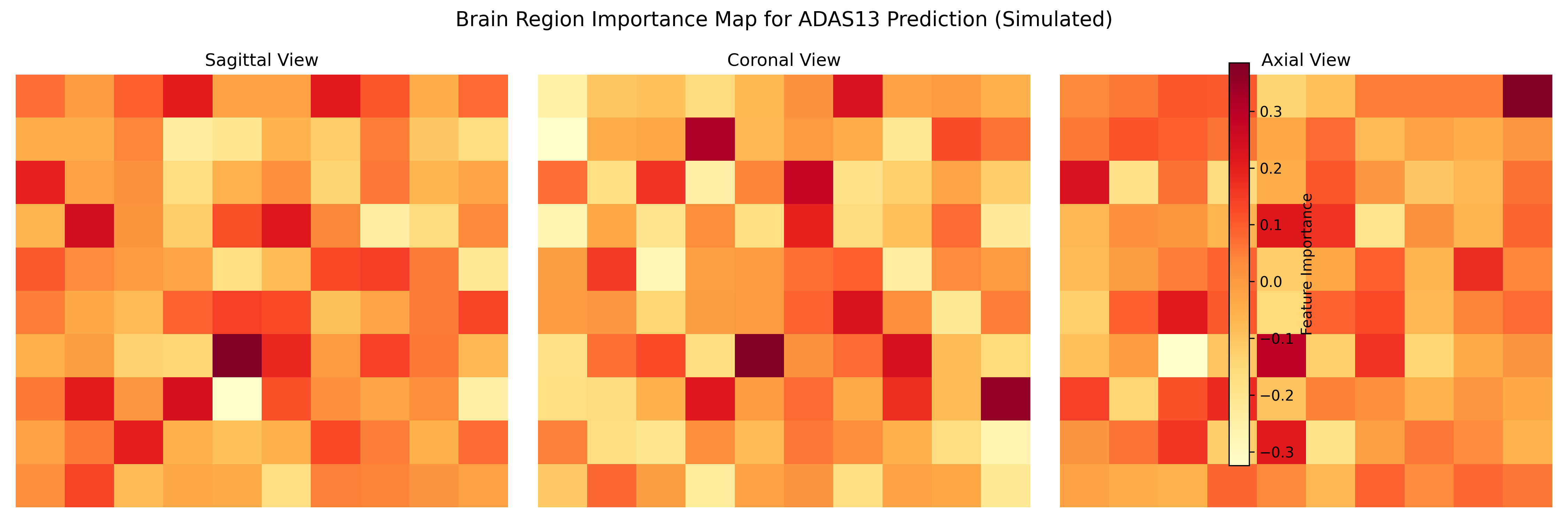}
    \caption{Spatial projection of the structural MRI feature importance scores onto approximate anatomical locations. The visualization highlights medial-temporal and ventricular regions as the leading drivers of ADAS13 next-visit prediction.}
    \label{fig:brain_map}
\end{figure}

To probe interpretability of the fusion stage, we trained a gradient-boosted surrogate over the structural MRI panel to attribute ADAS13 prediction to individual regions. Hippocampal and ventricular measurements emerged as the dominant contributors, in agreement with established AD imaging biomarkers, with mid-temporal and entorhinal regions providing secondary signal. Figure~\ref{fig:feature_importance} summarizes the ranked importances, and Figure~\ref{fig:brain_map} projects these scores onto approximate anatomical locations to convey the spatial pattern at a glance. This pattern supports the interpretability of the multimodal fusion component and aligns with the spatial priors that motivate the Bayesian low-rank tensor model described next.

\subsection{Imaging-Fusion Feasibility}
To assess the imaging-fusion component, we additionally evaluate a Bayesian low-rank tensor model on structural MRI data from ADNI. This analysis suggests that the tensor formulation can localize AD-sensitive regions while maintaining a parsimonious coefficient representation. Because TADPOLE primarily provides summarized imaging measures rather than voxel-level images, the ADNI analysis represents a component-level feasibility study rather than an end-to-end result within the full PCD-DT pipeline.

\subsection{Current Limitations}
The present results establish feasibility but do not yet constitute full predictive validation of the complete digital twin. While the next-visit ablation provides initial end-to-end evidence on continuous targets, several gaps remain. We do not yet report posterior uncertainty calibration, multi-step forecasting beyond a single visit, or robustness under systematic missingness. The diagnosis classification reached high test accuracy but produced unstable AUC under the current split because of limited class diversity in held-out folds, indicating the need for stratified and patient-disjoint evaluation. The LSTM model in this study also serves as a deterministic placeholder for the deeper variational state-space formulation, and replacing it with the full latent dynamics model is a near-term priority. Addressing these gaps is a key direction for future work.

\section{Conclusion}
We presented PCD-DT, a multimodal and uncertainty-aware framework for personalized digital twins in cognitive decline. The framework couples latent temporal dynamics with parsimonious multimodal fusion and uncertainty-aware updating. Preliminary analyses indicate that TADPOLE contains strong longitudinal signal for individualized modeling and that Bayesian tensor methods are promising for imaging integration. The multimodal next-visit ablation further shows that fusing cognitive scores with structural MRI yields the lowest standardized RMSE for both ADAS13 and ventricle volume and outperforms a LOCF baseline, providing concrete predictive evidence for the temporal and fusion components. A key next step is to extend this evaluation with the full variational state-space model, posterior uncertainty calibration, and longer-horizon trajectory forecasting on held-out patients. Overall, this work establishes a principled and extensible foundation for next-generation digital twin systems in neurodegenerative disease.

\bibliographystyle{IEEEtran}
\bibliography{_paper}

\end{document}